\documentclass{article}
\usepackage{cite}
\usepackage[utf8]{inputenc}
\usepackage[english]{babel}
\usepackage{graphicx}
\usepackage{amsmath}
\usepackage{float}
\usepackage{subcaption}
\usepackage{mlspconf}

\title{Understanding Patch-Based Learning by Explaining Predictions}

\name{Christopher Anders$^1$, Gr{\'e}goire Montavon$^1$, Wojciech Samek$^2$, Klaus-Robert M{\"u}ller$^{1,3,4}$\thanks{This work was supported by the German Ministry for Education and Research as Berlin Big Data Center under Grant 01IS14013A and by the Institute for Information \& Communications Technology Promotion and funded by the Korea government (MSIT) (No.\ 2017-0-01779 and No.\ 2017-0-00451).}}
\address{$^1$Dept.\ of Computer Science,Technische Universit{\"a}t Berlin, 10587 Berlin, Germany\\
$^2$Dept.\ of Video Coding \& Analytics, Fraunhofer Heinrich Hertz Institute, 10587 Berlin, Germany\\ 
$^3$Dept.\ of Brain \& Cognitive Engineering, Korea University, Seoul 136-713, South Korea\\
$^4$Max Planck Institute for Informatics, Saarbr{\"u}cken 66123, Germany}

\begin{document}

\maketitle

\begin{abstract}
Deep networks are able to learn highly predictive models of video data.
Due to video length, a common strategy is to train them on small video snippets.
We apply the deep Taylor / LRP technique to understand the deep network's classification decisions, and identify a ``border effect'': a tendency of the classifier to look mainly at the bordering frames of the input.
This effect relates to the step size used to build the video snippet, which we can then tune in order to improve the classifier's accuracy without retraining the model.
To our knowledge, this is the the first work to apply the deep Taylor / LRP technique on any video analyzing neural network.
\end{abstract}


\section{Introduction}
\label{sec:intro}
Deep neural networks have set new standards of performance in many machine learning areas such as image classification \cite{krizhevsky2012imagenet,szegedy2015going}, speech recognition \cite{graves2013speech,oord2016wavenet}, or video analysis \cite{ji20133d,karpathy2014large}.
For applications where the input signal is very large in time or space, it has been a common practice to train the model on small patches or snippets of that signal \cite{ji20133d,hou2016patch,BosTIP17}.
This strategy reduces the number of input variables to be processed by the network and thus, allows to extract the problem's nonlinearities more quickly by performing more training iterations.

An underlying assumption of patch- or snippets-based training is the locality of the label information.
This assumption is often violated in practice:
For example, discriminative information may only be contained in long-term interactions \cite{graves2013speech,yue2015beyond,oord2016wavenet} or only reside at specific time steps (e.g.\ when a particular action occurs).
Since such label noise makes the training more difficult \cite{reed2014training}, recent work investigated ways to cope with this problem, e.g., attention mechanisms \cite{sharma2015action} or weighted patch aggregation \cite{BosTIP17}.

This paper aims to investigate patch- or snippets-based learning from another perspective, namely by analyzing the properties of a model trained with this specific learning procedure.
One way to study the properties of a model is to perform introspection into how the model predicts, for example, by explaining its predictions in terms of input variables.
Such explanations can now be robustly obtained for a wide range of convolution-type or general deep neural networks \cite{zeiler2014visualizing,springenberg2014striving, yosinski2015understanding, bach2015pixel,ribeiro2016should,montavon2017explaining}.

In this work we analyzed a convolutional neural network \cite{tran2015learning} trained for human action recognition on the Sports1M data \cite{karpathy2014large}
using the deep Taylor / LRP decomposition technique \cite{montavon2017explaining}.
We first show that this explanation technique reliably captures class-relevant information from videos.
We then test how snippets-based training affects the prediction strategy of the network and identify two effects induced by this training procedure.
The ``border effect'' describes the observation that the prediction predominantly looks at the border of the signal given as input to compensate for a small snippet size,
whereas the ``lookahead effect'' describes the observation that the model learns to ignore the first few frames of the video and assign more relevance to the later ones.
Finally we demonstrate that the insights obtained by explaining predictions can be directly (i.e. without retraining) used to increase the prediction accuracy of the classifier.

\label{ssec:related}
While a different approach for human action recognition has been analyzed before \cite{srinivasan2017interpretable} using the LRP framework \cite{bach2015pixel}, to our knowledge this work is the first to analyze any neural network for video classification using deep Taylor decomposition \cite{montavon2017explaining}.
In a recent work, voxel explanations of 3D-CNNs \cite{yang2018visual} have been produced using different explanation frameworks \cite{zhou2016learning,selvaraju2016grad}.
Further research has been done on the interpretation \cite{donahue2015long}, description \cite{zhang2016automatic} and segmentation of videos \cite{pohlen2017full}.
Outside the field of machine learning, some work has been done on saliency detection in videos \cite{hu2013spatiotemporal,li2015spatiotemporal} .

\begin{figure*}[t!]
    \centering
    \begin{subfigure}[b]{0.9\linewidth}
    \parbox{0.166\textwidth}{\footnotesize \centering frame 1}%
    \parbox{0.166\textwidth}{\footnotesize \centering frame 4}%
    \parbox{0.166\textwidth}{\footnotesize \centering frame 7}%
    \parbox{0.166\textwidth}{\footnotesize \centering frame 10}%
    \parbox{0.166\textwidth}{\footnotesize \centering frame 13}%
    \parbox{0.166\textwidth}{\footnotesize \centering frame 16}\\[+2px]
        \frame{\includegraphics[width=\linewidth]{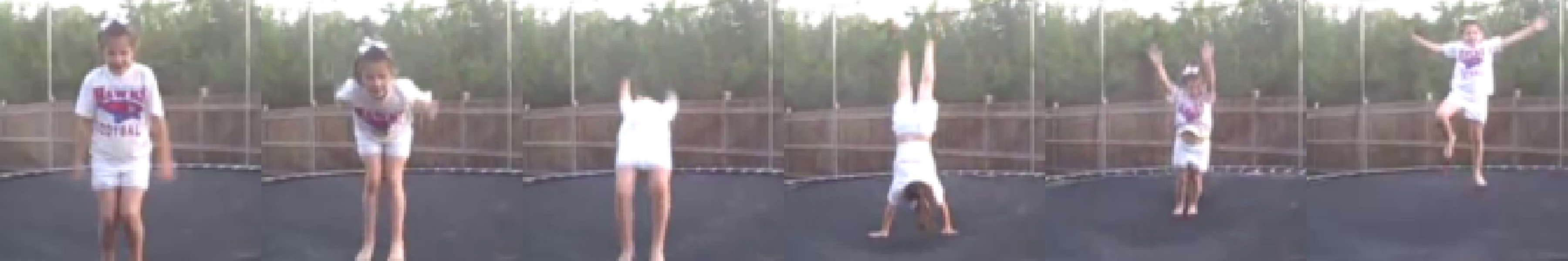}}\vskip -1mm
        \caption{Data}
        \label{fig:tumbling_data}
    \end{subfigure}
    \hfill
    \begin{subfigure}[b]{0.9\linewidth}
        \vskip +1mm
        \frame{\includegraphics[width=\linewidth]{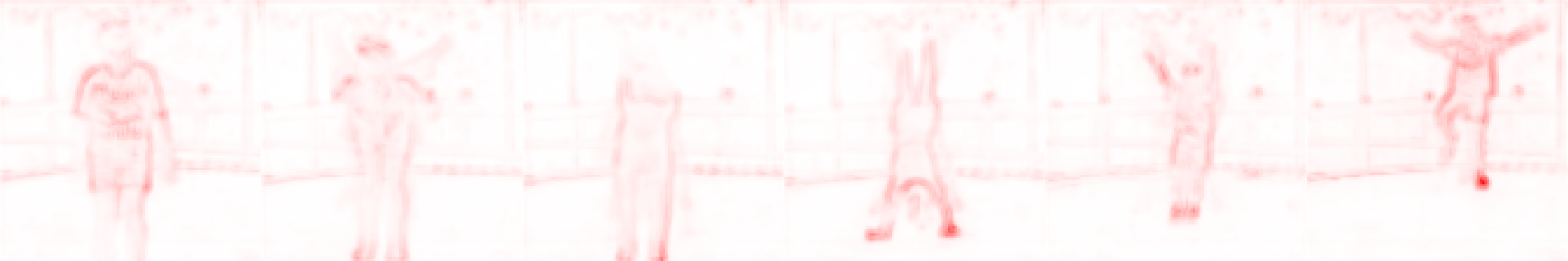}}\vskip -1mm
        \caption{Deep Taylor Decomposition (DTD)}
        \label{fig:tumbling_hm}
    \end{subfigure}
    \hfill
    \begin{subfigure}[b]{0.45\linewidth}
        \vskip +1mm
        \frame{\includegraphics[width=\linewidth]{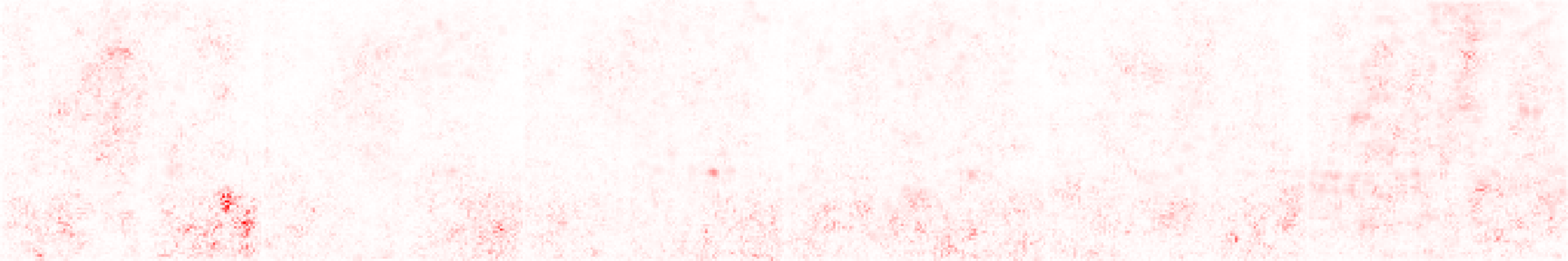}}\vskip -1mm
        \caption{Sensitivity Analysis}
        \label{fig:tumbling_grad}
    \end{subfigure}
    \begin{subfigure}[b]{0.45\linewidth}
        \vskip +1mm
        \frame{\includegraphics[width=\linewidth]{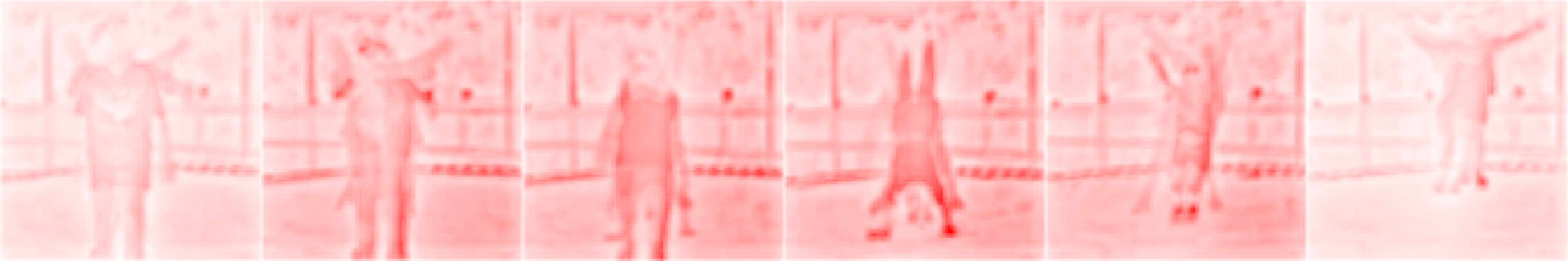}}\vskip -1mm
        \caption{DTD on untrained model}
        \label{fig:tumbling_rand}
    \end{subfigure}
    \caption{Example of a video along with the DTD explanation of this video belonging to the class `Tumbling'. High relevance scores are shown in red.}
    \label{fig:tumbling}
\end{figure*}

\section{Explaining the classifier's predictions}
\label{sec:methods}
In this paper, we use the deep Taylor / LRP decomposition technique \cite{montavon2017explaining} to produce explanations.
We give a brief textual description of the method, along with connections to previous work.
The method performs a sum-decomposition of the function value $f(x)$ in terms of input variables
\begin{align}
f(x) = \sum_{p,t} R_{p,t}
\label{eq:decomp}
\end{align}
\cite{poulin2006visual}, where $R_{p,t}$ is the relevance of pixel $p$ in frame $t$.
These scores are obtained by propagating the output $f(x)$ backwards in the network until the input variables are reached.
The propagation procedure satisfies a conservation principle \cite{landecker2013interpreting,bach2015pixel}, where each neuron redistributes to the lower-layer as much as it has received from the higher layer.
Let $i,j$ be neurons of adjacent layers. Let $a_i$ be the activation of neuron $i$ and $w_{ij}$ be the weight that connects it to neuron $j$.
In hidden layers, the redistribution is in proportion to the positive contribution of the input activations
$R_{i \leftarrow j} \propto a_i w_{ij}^+$
of each neuron \cite{bach2015pixel,montavon2017explaining}.
In pooling layers, the redistribution is in proportion to the activations $a_i$ inside the pool \cite{montavon2017explaining}.
For the first convolutional layer we redistribute in proportion to the signed contributions plus some additive term
$R_{i \leftarrow j} \propto a_i w_{ij}- l_i w_{ij}^+ - h_i w_{ij}^-$
where $l_i$ and $h_i$ are the min and max pixel value \cite{montavon2017explaining}.

Another popular explanation technique is sensitivity analysis \cite{gevrey2003review,simonyan2013deep}, which computes importance scores as e.g.
\begin{align}
S_{p,t} = \Big(\frac{df}{dx_{p,t}}\Big)^2 \text{.}
\label{eq:sa}
\end{align}
We note that this analysis can be interpreted as performing a sum-decomposition of the squared gradient norm ($\|\nabla f\|^2 = \sum_{p,t} S_{p,t} $), and is thus closer to an explanation of the function's variation.
We refer the reader for a comparison of different explanation methods to \cite{SamTNNLS16, MonDSP17}.

\section{Experiments}
\label{sec:experiments}
We use the 3-dimensional convolutional neural network architecture C3D as described in \cite{tran2015learning}, trained on the Sports-1M data set, which consists of roughly 1 million sports videos from YouTube with 487 classes \cite{karpathy2014large}.
Videos are pre-processed by spatially resizing to $128 \times 171$ pixels and then center-cropping to $121 \times 121$ pixels.
We take video snippets at particular offsets, composed of $16$ frames each.

We explain predictions for $1000$ videos from the test set of Sports-1M using deep Taylor decomposition \cite{montavon2017explaining}.
Additional explanations are given for the same 3-dimensional convolutional neural network architecture untrained as well as using gradient-based sensitivity analysis \cite{gevrey2003review,simonyan2013deep} for comparison.

\subsection{Heatmap Analysis}
\label{ssec:quality}
To get a first impression of the prediction, we take a look at the individual explanation of one specific video snippet.
In Fig.\ \ref{fig:tumbling}, we show an exemplary video and the deep Taylor decomposition (DTD) for the predicted class ``Tumbling''. The hands are identified as relevant, especially when the latter are touching the trampoline, which is characteristic of that class. Other parts of the image such as the trees in the background are not highlighted and therefore found to be non-relevant. The DTD analysis is also less noisy and more focused on the class-relevant features than sensitivity analysis (Fig.\ \ref{fig:tumbling_grad}).

An interesting observation that can be made is that the training procedure tends to make the relevance converge from the center of the video sequence to its borders as evidenced by the difference between DTD and the same analysis performed on an untrained network (Fig.\ \ref{fig:tumbling_rand}). This border effect will be studied quantitatively in Section \ref{ssec:border}. The initial focus on the center of the sequence is due to these frames being more densely connected to the output.

\begin{figure}[t!]
    \centering
    \begin{subfigure}[b]{0.9\linewidth}
     \parbox{0.166\textwidth}{\footnotesize \centering frame 1}%
     \parbox{0.166\textwidth}{\footnotesize \centering frame 4}%
     \parbox{0.166\textwidth}{\footnotesize \centering frame 7}%
     \parbox{0.166\textwidth}{\footnotesize \centering frame 10}%
     \parbox{0.166\textwidth}{\footnotesize \centering frame 13}%
     \parbox{0.166\textwidth}{\footnotesize \centering frame 16}\\[+2px]
         \frame{\includegraphics[width=\linewidth]{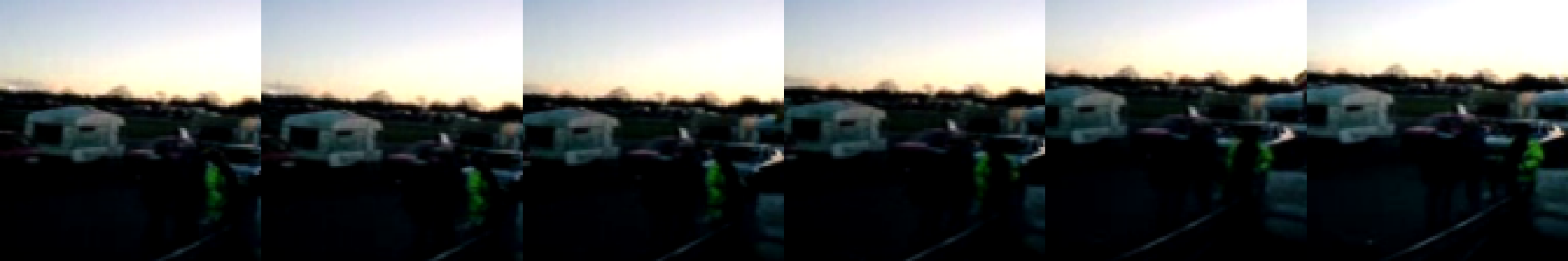}} \vskip 0.5mm
         \frame{\includegraphics[width=\linewidth]{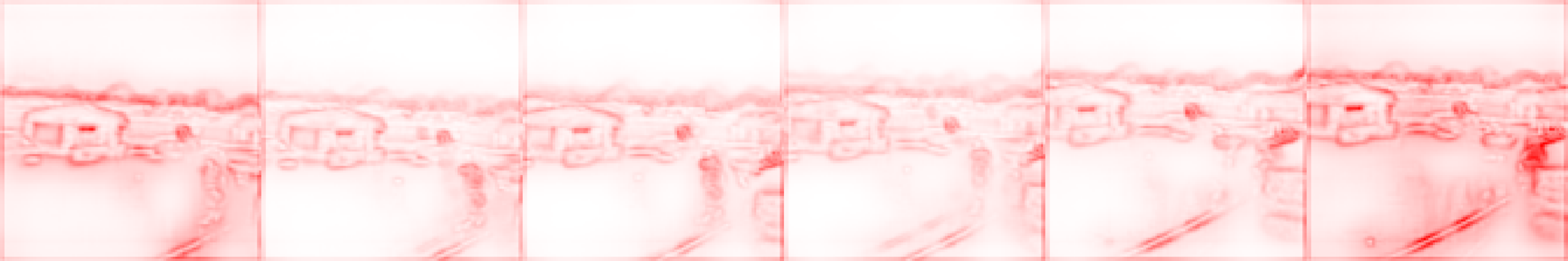}} \vskip 0.5mm
         \frame{\includegraphics[width=\linewidth]{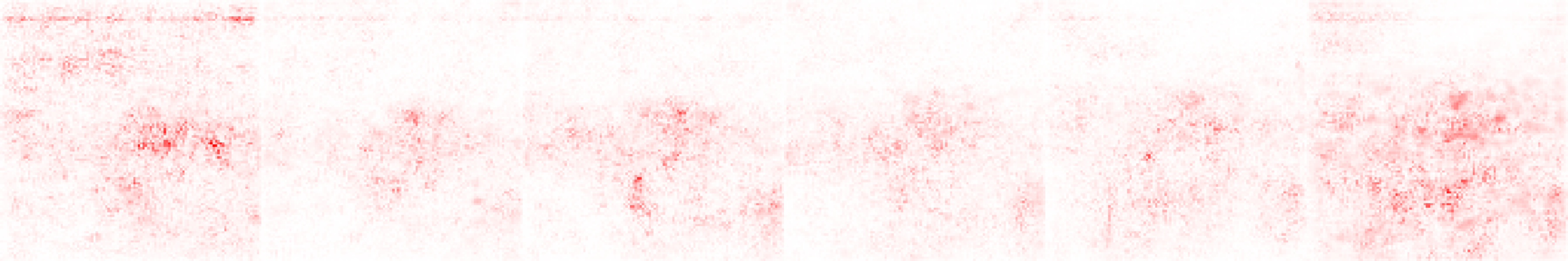}} \vskip 0.5mm
         \caption{Demolition Derby (Carting)}
         \label{fig:bulk2}
    \end{subfigure}
    \\[+3mm]
    \begin{subfigure}[b]{0.9\linewidth}
     \parbox{0.166\textwidth}{\footnotesize \centering frame 1}%
     \parbox{0.166\textwidth}{\footnotesize \centering frame 4}%
     \parbox{0.166\textwidth}{\footnotesize \centering frame 7}%
     \parbox{0.166\textwidth}{\footnotesize \centering frame 10}%
     \parbox{0.166\textwidth}{\footnotesize \centering frame 13}%
     \parbox{0.166\textwidth}{\footnotesize \centering frame 16}\\[+2px]
         \frame{\includegraphics[width=\linewidth]{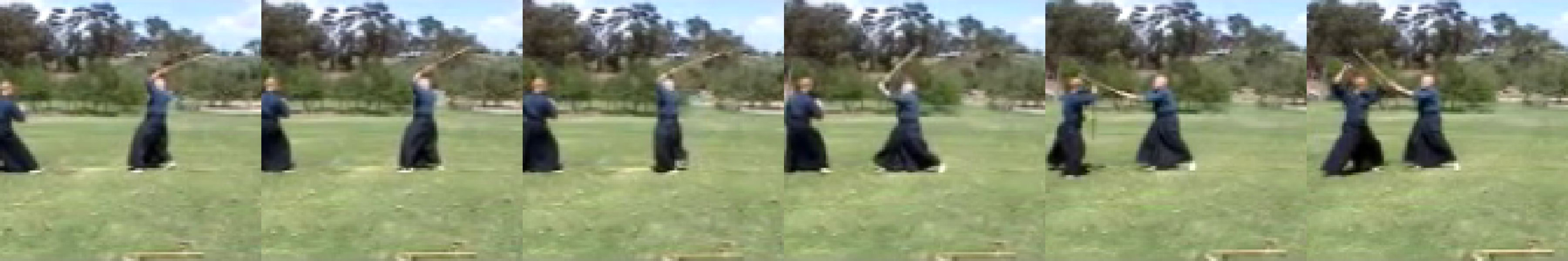}} \vskip 0.5mm
         \frame{\includegraphics[width=\linewidth]{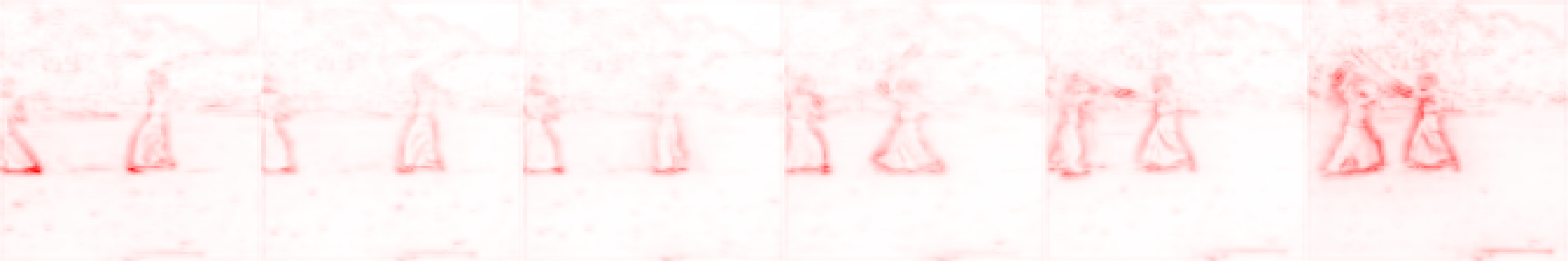}} \vskip 0.5mm
         \frame{\includegraphics[width=\linewidth]{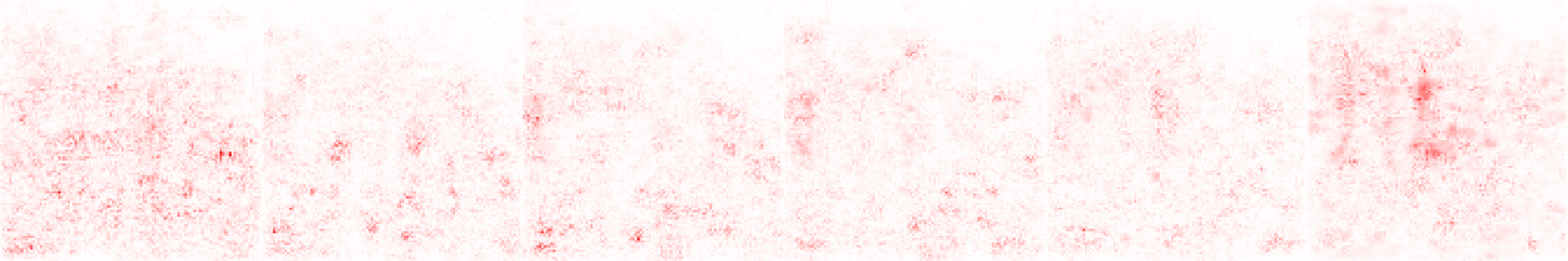}} \vskip 0.5mm
         \caption{J{\={o}}d{\={o}} (Kenjutsu)}
         \label{fig:bulk4}
    \end{subfigure}
    \\[+3mm]
    \begin{subfigure}[b]{0.9\linewidth}
     \parbox{0.166\textwidth}{\footnotesize \centering frame 1}%
     \parbox{0.166\textwidth}{\footnotesize \centering frame 4}%
     \parbox{0.166\textwidth}{\footnotesize \centering frame 7}%
     \parbox{0.166\textwidth}{\footnotesize \centering frame 10}%
     \parbox{0.166\textwidth}{\footnotesize \centering frame 13}%
     \parbox{0.166\textwidth}{\footnotesize \centering frame 16}\\[+2px]
         \frame{\includegraphics[width=\linewidth]{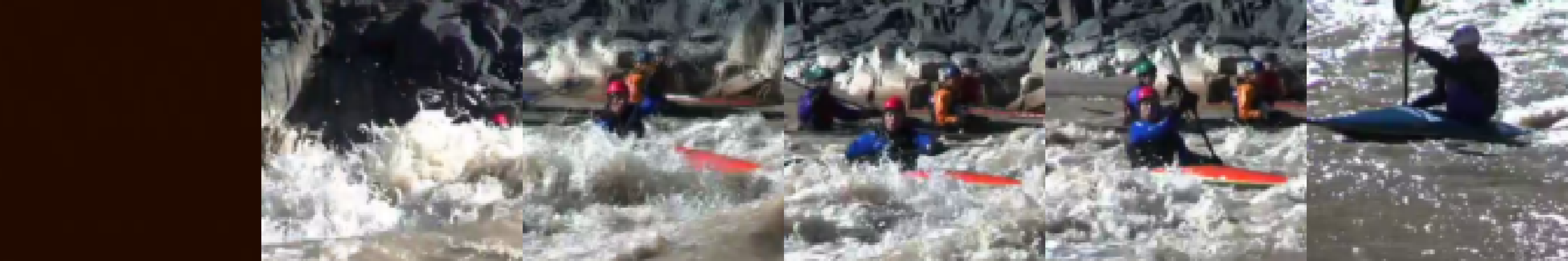}} \vskip 0.5mm
         \frame{\includegraphics[width=\linewidth]{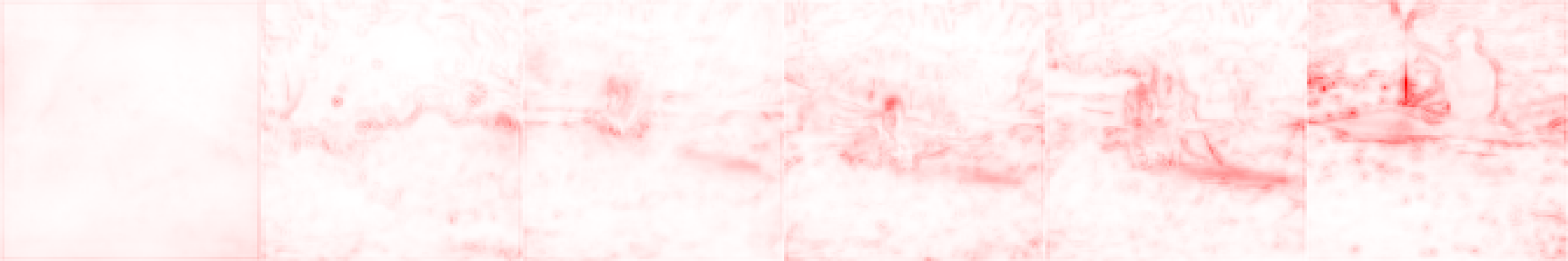}} \vskip 0.5mm
         \frame{\includegraphics[width=\linewidth]{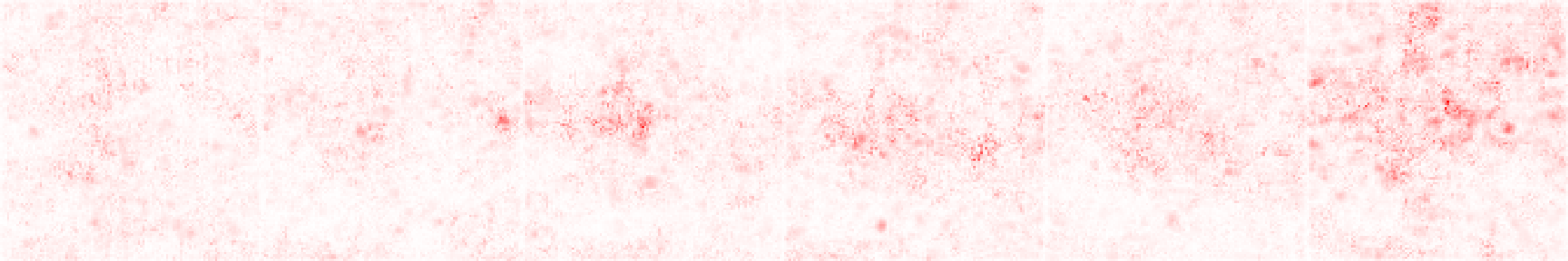}} \vskip 0.5mm
         \caption{Whitewater Kayaking (Whitewater Kayaking)}
         \label{fig:bulk6}
    \end{subfigure}
    \\[+3mm]
    \begin{subfigure}[b]{0.9\linewidth}
    \parbox{0.166\textwidth}{\footnotesize \centering frame 1}%
    \parbox{0.166\textwidth}{\footnotesize \centering frame 4}%
    \parbox{0.166\textwidth}{\footnotesize \centering frame 7}%
    \parbox{0.166\textwidth}{\footnotesize \centering frame 10}%
    \parbox{0.166\textwidth}{\footnotesize \centering frame 13}%
    \parbox{0.166\textwidth}{\footnotesize \centering frame 16}\\[+2px]
        \frame{\includegraphics[width=\linewidth]{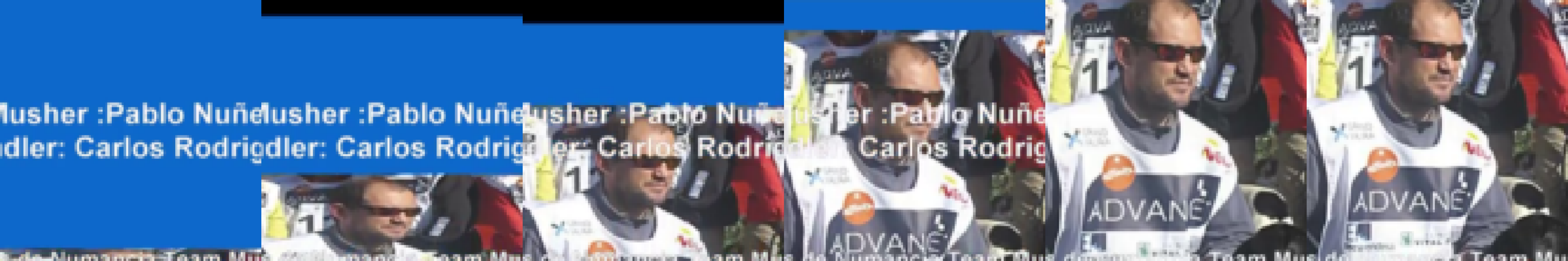}} \vskip 0.5mm
        \frame{\includegraphics[width=\linewidth]{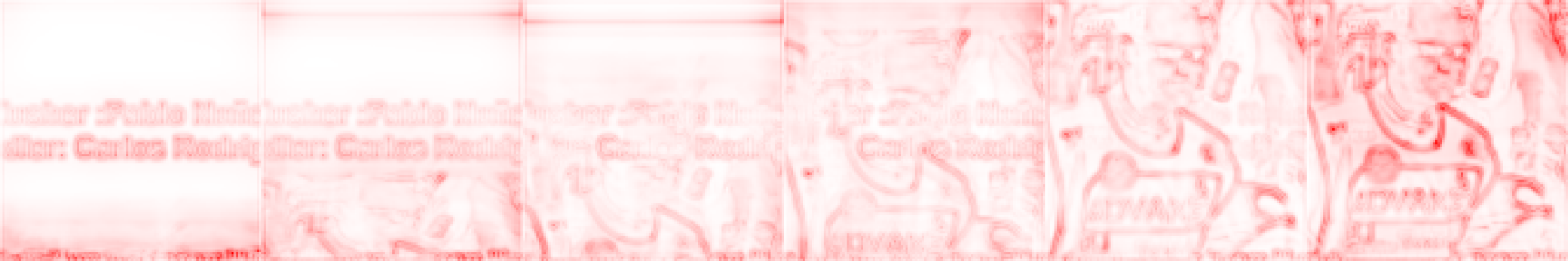}} \vskip 0.5mm
        \frame{\includegraphics[width=\linewidth]{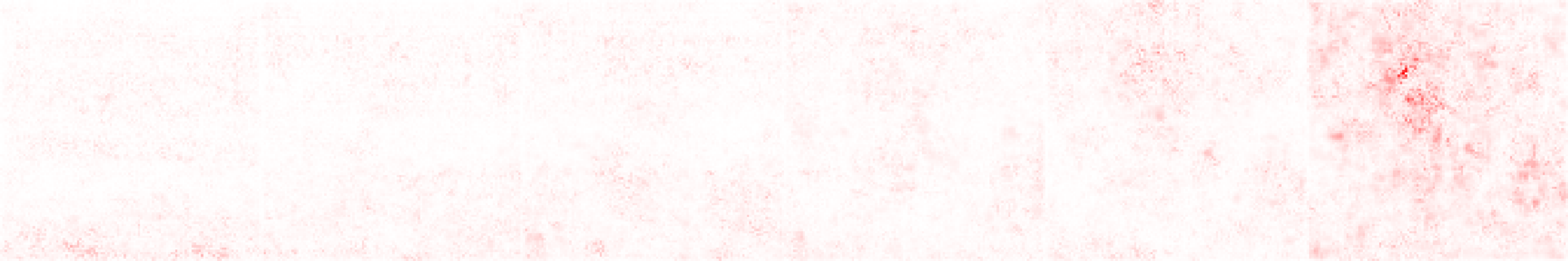}} \vskip 0.5mm
        \caption{Mushing (Gridiron Football)}
        \label{fig:bulk7}
    \end{subfigure}
    \caption{Examples of videos belonging to different classes. For each example from top to bottom: Input Video, Deep Taylor Decomposition, Sensitivity Analysis. Captions are the true label followed by the predicted label in curved brackets.}
    \label{fig:bulk}
\end{figure}

Additional examples of different videos are shown in Fig.\ \ref{fig:bulk}. In particular, the aforementioned oberservation of higher relevance towards the videos' borders is more clearly visible in Figs.\ \ref{fig:bulk2} and \ref{fig:bulk4}. Furthermore, we can also observe that the final frames receive more relevance than any other ones in Figs.\ \ref{fig:bulk2}, \ref{fig:bulk6} and \ref{fig:bulk7}. This lookahead effect will also be studied quantitatively in Section \ref{ssec:border}.

\subsection{Quantifying Border and Lookahead Effects}
\label{ssec:border}

\begin{figure}[t!]
    \centering
    \begin{subfigure}[b]{\linewidth}
        \includegraphics[width=\linewidth]{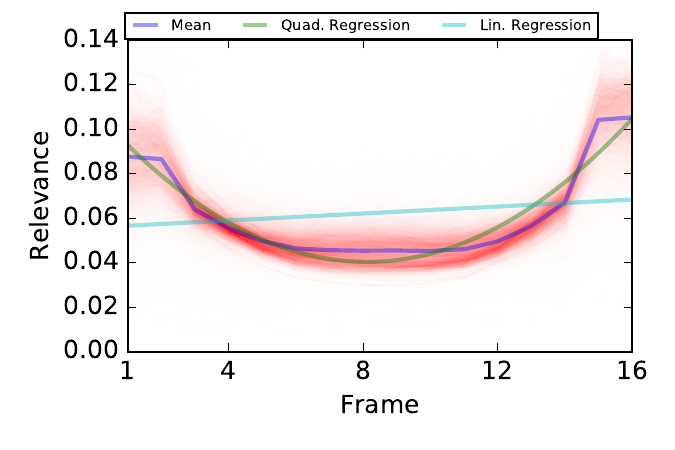}\vskip -3mm
        \caption{Deep Taylor Decomposition}
        \label{fig:distr_temp}
    \end{subfigure}
    \hfill
    \begin{subfigure}[b]{0.49\linewidth}
        \includegraphics[width=\linewidth]{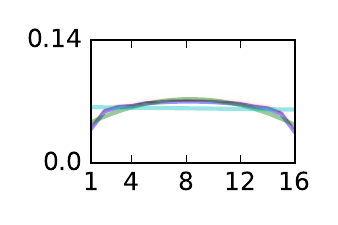}\vskip -3mm
        \caption{DTD on untrained model}
        \label{fig:distr_random}
    \end{subfigure}
    \begin{subfigure}[b]{0.49\linewidth}
        \includegraphics[width=\linewidth]{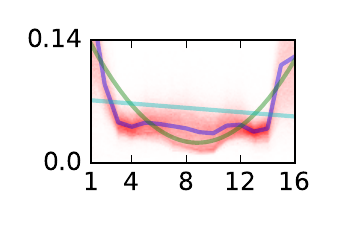}\vskip -3mm
        \caption{Sensitivity Analysis}
        \label{fig:distr_grad}
    \end{subfigure}
    \caption{Relevance share $(P_t)_t$. Red color shows these vectors for a large number of videos. Lines show the mean relevance share and polynomial fits.}
    \label{fig:dimrel}
\end{figure}

To refine the intuition developed in Section \ref{ssec:quality} about the presence of a border and lookahead effect, we produce DTD explanations for a large number of videos and analyze their average properties.
Because the border effect occurs in the temporal domain, we only focus on the temporal axis of explanations $R_{p,t}$ by defining frame-wise explanation
\begin{align*}
R_t = \sum_{p} R_{p,t}
\end{align*}
From these relevance scores, we can define a vector $(P_t)_t$ where $P_t = R_t / \sum_t R_t$ is the share of relevance at time $t$.
Since our input videos contain $16$ frames, this vector has size $16$, which we can visualize in a plot.
Results are shown in Fig.\ \ref{fig:dimrel}.
The red pattern represents the distribution of these $16$-dimensional vectors, for which we can compute an average over the data set (blue line).
Results are also compared to sensitivity analysis, as well as DTD on the untrained model.

Results confirm our previous observations of higher relevance in the bordering frames. Note that DTD and sensitivity analysis (Fig.\ \ref{fig:distr_grad}) produce consistent results with respect to the border effect. We can further verify that this effect is not due to an architecture-related artifact, by peforming the same DTD analysis on the untrained model (Fig.\ \ref{fig:distr_random}): The border effect is present only for the trained model. For the untrained model, relevance at the border is instead lower compared to other frames.
The additional lookahead effect can be observed from this analysis where the relevance is slightly higher for the last frame as opposed to the first frame.

In order to determine the strength of the border and lookahead effects, we need a quantitative measure for them.
We propose to capture these effects by fitting vectors $(P_t)_t$ using simple quadratic regression. More specifically, we consider the quadratic model
\begin{align}
q(t) = B \cdot t^2 + C \cdot t + D
\label{eq:quad}
\end{align}
and fit the coefficients $B, C, D$ to minimize the least square error
\begin{align*}
\sum_t \| E[ P_t ] - q(t)\|^2 \text{,}
\end{align*}
where $E[\cdot]$ is the expectation over the Sports-1M test set.
The strength of the border effect is captured by the variable $B$.
Similarly, to capture the lookahead effect, we fit a linear model
\begin{align}
l(t) = L \cdot t + A
\label{eq:lin}
\end{align}
using similar least squares objective, and identify the lookahead strength by the parameter $L$. Fitted models $q(t)$ and $l(t)$ are shown as green and cyan lines in Fig.\ \ref{fig:dimrel}.

\begin{table}[h!]
\begin{center} \small
\begin{tabular}{c|ccc}
  & DTD & SA & DTD-u\\[+0.5mm]\hline
  &&&\\[-1mm]
$\boldsymbol{B}$      & $\bf 0.0010$ & $\bf 0.0018$  & $\bf -0.0005$  \\
$C$      & $-0.0168$ & $-0.0322$ & $0.0082$    \\
$D$      & $0.1085$ & $0.1661$  & $0.0389$   \\[+2mm]
$\boldsymbol{L}$      & $\bf 0.0007$  & $\bf -0.0012$ & $\bf -0.0002$   \\
$A$      & $0.0558$ & $0.0729$  & $0.0640$
\end{tabular}
\end{center}
\vskip -5mm
\caption{Parameters for fitted models $q(t)$ and $l(t)$ as in Eqs. \ref{eq:quad} and \ref{eq:lin}. Relevant coefficients are shown in bold.}
\label{tab:params}
\end{table}

These parameters are shown in Table \ref{tab:params} for the deep Taylor decomposition (DTD), sensitivity analysis (SA), and the  DTD on the untrained model (DTD-u).
Coefficients used for the analysis are shown in bold. We can observe that the border parameter $B$ is positive for both analyses performed on the trained model. The lookahead parameter however has varying signs depending on the choice of analysis. We will see later in Section \ref{ssec:lookahead} that this parameter is influenced by the offset of the input sequence.
\subsection{Border Effect and Step Size}
\label{ssec:stepsize}

\begin{figure}[t!]
  \centering
  \includegraphics[width=\linewidth]{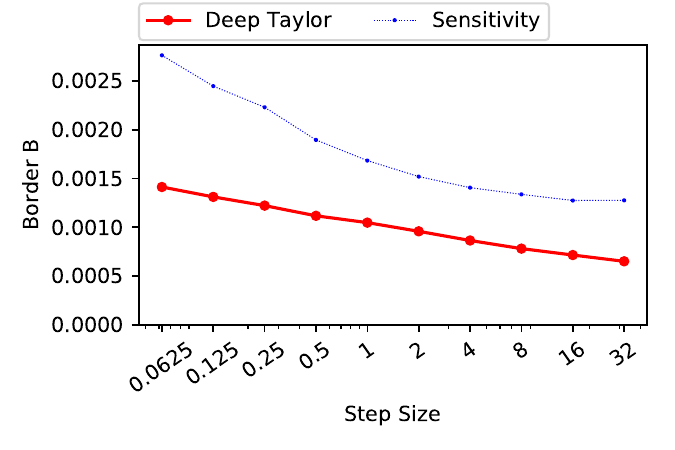}
  \vskip -5mm
  \caption{Border Parameter $B$ by Step Size (logarithmic scale)}
  \label{fig:step}
\end{figure}

The border effect can be intuitively understood as an attempt by the network to look beyond the sequence received as input. This suggests that upscaling the input sequence may reduce this effect as more context becomes available.
For example, Fig.\ \ref{fig:bulk2} is a static scene with barely any motion and shows, compared to other samples, more relevance at the border frames.
To test this, we will subsample videos with various step sizes.
We start with a step size of $\frac{1}{16}$, which is the same frame repeated 16 times.
We then double the step size repeatedly until we reach a value of 32.
At each step size, we apply DTD as well a sensitivity analysis. Note that the model is left untouched.
The border parameter $B$ for each step size is given in Fig.\ \ref{fig:step}.
For low step sizes, the border effect is strong. As the step size increases, the border effect is reduced, thus confirming the above intuition.

\subsection{Lookahead Effect and Offset}
\label{ssec:lookahead}

\begin{figure}[t!]
  \centering
  \includegraphics[width=\linewidth]{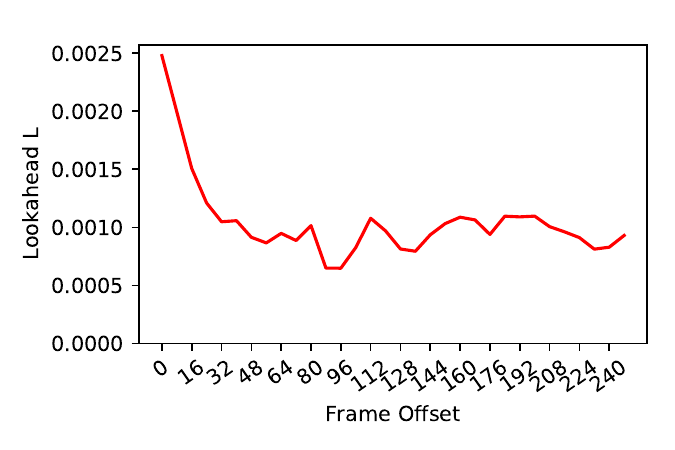}
  \vskip -5mm
  \caption{Lookahead Parameter $L$ by Intra-Video Frame Offset.}
  \label{fig:offset}
\end{figure}

The lookahead effect is the tendency of the network to look predominantly at the end of the sequence. We would like to test whether this effect occurs at every position in the video or mainly at the beginning.
One of our suspicions is, that many videos start with some opening screen, where the title of the video, authors etc. are introduced.
It would seem natural that the model ignores the first few frames of the video and assigns more relevance to the later frames.
An example for such a video snippet is shown in Fig.\ \ref{fig:bulk7}.
We start by taking the input sequence at the beginning of the video, then, we slide the window by 8-framed offsets until we reach an offset of 256.
The results are shown in Fig.\ \ref{fig:offset}.
We observe that for offsets 0, 8 and 16, the lookahead parameter is high compared to other offsets, and becomes low and constant for larger offsets. This behavior for small offsets supports the hypothesis of non-informative content at the beginning of the video.

\begin{figure}[t!]
  \centering
  \includegraphics[width=\linewidth]{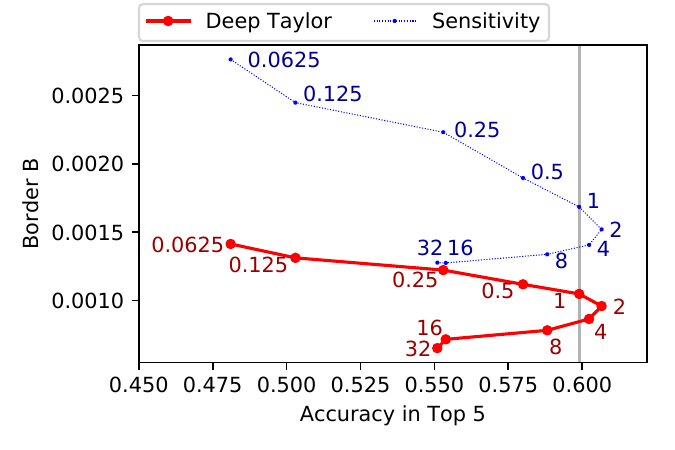}
  \vskip -5mm
  \caption{Border parameter $B$ by accuracy in top 5 along step size. The grey bar indicates the baseline accuracy.}
  \label{fig:accuracy}
\end{figure}

\subsection{Step Size and Model Accuracy}
\label{ssec:accuracy}

As a final experiment, we look at how the step size not only controls the border effect, but also the model's classification accuracy. In particular, we test whether we can improve the classifier accuracy by simply choosing a step size different from the training data, without retraining the model.
We use the previously defined step sizes and plot in Fig.\ \ref{fig:accuracy} the resulting border parameter in correspondence to the produced classification accuracy. (The measure of accuracy is the membership of the true label to the top five predictions.)
A low step size produces few correct predictions. Performance slowly increases until the highest accuracy is reached at a step size of $2$, about $1\%$ above the baseline accuracy of $60\%$.
After that, accuracy drops again until a step size of $16$. A key observation here is that the optimal step size is different from the step size $1$ used for training the model.
Thus, the classification accuracy was improved at no cost.

\section{Conclusion}
\label{sec:conclusion}
In this work, we have explained the reasoning of a highly predictive video neural network trained on a sports classification task.
For this, we have used the recently proposed deep Taylor / LRP framework, which allowed us to robustly identify which frames in the video and which pixels of each frame are relevant for prediction.
The method was able to correctly identify video features specific to certain sports. In addition, the analysis has also revealed systematic imbalances in the way relevance is distributed in the temporal domain.
These imbalances, that we called "border effect" and "lookahead effect", can be understood as an attempt by the network to look beyond the sequence it receives as input.
Based on the result of this analysis, we then explored how transforming the input data reduces/increases these imbalances.
In particular, downsampling the data was shown to reduce the border effect, and also to bring a small increase in classification accuracy (Fig. \ref{fig:accuracy}), without actually retraining the model.

\bibliographystyle{abbrv}
\bibliography{border}

\end{document}